\documentclass[runningheads]{llncs}
\usepackage{amsmath}
\usepackage{amssymb}
\usepackage{geometry}
\usepackage[backend=biber]{biblatex}
\usepackage[T1]{fontenc}
\usepackage{hyperref}
\usepackage{orcidlink}
\usepackage{tikz}
\usepackage{graphicx}
\usetikzlibrary{calc,backgrounds,decorations.pathmorphing, fit}
\tikzset{stroke/.style={draw=black}}
\title{Convolutional Model Trees}
\author{ William Armstrong\inst{1}\orcidID{0009-0004-2657-2995}\and 
	Hongyi Li \inst{2}\orcidID{0000-0003-3585-7365}
	 \and Jun Xu \inst{2}\orcidID{0000-0002-2934-4814}}
\authorrunning{W. W. Armstrong et al.}
\institute{University of Alberta, Edmonton, Canada \and
	Harbin Institute of Technology, Shenzhen, China}
\date{Revised January 21, 2026 \copyright{(CC BY 4.0)}}
\begin{document}
\maketitle
\begin{abstract}
A method for creating a forest of model trees to fit samples of a function defined on images is described in several steps:  down-sampling the images, determining a model tree's hyperplanes that partition the image space into blocks, applying convolutions to the hyperplanes and leaf-block functions to process images as though the images themselves had undergone convolution, and creating forests of model trees to increase accuracy and achieve a smooth fit. One-to-one correspondences among sensors of images, coefficients of hyperplanes and coefficients of leaf-block functions (excluding constant terms) support a method to create new trees to process large distortions of input images, such as arbitrary rotations or changes of perspective. This is done by distributing the values of coefficients of moved pixels to coefficient locations near the corresponding pixel destinations. A theoretical method for smoothing forest outputs to produce a continuously differentiable approximation is described and a training procedure is proved to converge.
\end{abstract}
\section{Introduction}
Regression from visual data is typically addressed using convolutional neural nets (CNNs) \cite{syamil2020cnn}. Although they have been very successful, their computation of an output from a given input using a number of weights and both linear and non-linear functions is hard to interpret. CNNs can, indeed must, compute output values over large parts of the domain even where guidance by samples is lacking.  On the other hand, model trees, which combine decision tree partitioning with linear regression models on the blocks of the partition, offer a more interpretable alternative, but they depend heavily on having sufficient data. One application of convolutional model trees (CMTs) is "distilling" \cite{hinton2015distilling,frosst2017distilling}, where the inputs and outputs of a CNN or back-propagation neural network are used to generate samples for a CMT which has appropriate HPs and leaf-block functions, with convolutions done in the tree if necessary.

Given a trained CMT, it is possible to create many new CMTs without training by shifting parts of the coefficient values of the trained CMT to new locations. The new CMTs can each handle images distorted in a corresponding way. This is like data augmentation but does not need any training.

The space of all possible images created using a given set of sensors, the \emph{image space}, is an axis-aligned hyper-rectangle (HR) where each axis contains the range of values of one sensor of the capturing camera. The dimension of the HR can range from around one hundred to several millions, as in the case of a 4K device. To reduce the dimension of the image space, it undergoes pooling, i.e. averaging or taking the maximum of neighboring pixels' values to form "group-pixel" values. From nere on we shall refer to these  simply as pixels in a lower-dimensional image space.

To create a model tree, we start with a training set of probably intensity- and contrast-normalized images which have not undergone convolution. A least-squares fit of the image samples with the desired output attached is computed.  The initial HR is recursively split into ever smaller convex blocks by hyperplanes (HPs). The details of how to use the least-squares fit to get HP coefficients and how to do the convolution on the coefficients will be described below. After each split, a least-squares fit of the samples in each of the two child blocks is computed to continue splitting. When the fitting function on a block approximates the block's sample data with less than a specified RMS error, the block does not split further but becomes a leaf-block of the CMT. It retains the HP coefficients after convolution which it would have used to split the block as its leaf-function. If the input images have undergone convolution, it is not done in the CMT.

The coefficients of HPs and leaf-functions (except the constant terms) can be thought of as arranged in a grid in one-to-one correspondence with the pixels of the images. Each pixel corresponds to a coefficient giving the rate of change of the output variable with respect to that pixel's intensity in the linear fit. As each HP was created, its coefficients were (optionally) subjected to convolution, say by a circularly symmetric kernel. An approximation of all samples formed by combining all the leaf-functions together is generally discontinuous, however we shall show theoretically that appropriately weighted outputs of several CMTs in a "forest" can be averaged to obtain a continuously differentiable ($C^{1}$) function.

For our purpose, we use a circularly symmetric kernel reflecting the fact that this represents pixel or coefficient locations at a certain distance from each other. Convolution enables HPs to better respond to slight movements, within the radius of the kernel, of images or parts of images used in training. Very importantly, capturing this convolution on the HPs costs no extra computing time after training. It is well-known that any self-adjoint real-valued kernel $K(x, y) = K(-x, -y)$ keeps the inner products equal whether one convolves either the images or the coefficients (but not both).

Algorithms such as M5 \cite{quinlan1992m5}, M5prime \cite{jekabsons2011m5primelab} and PILOT \cite{ raymaekers2024pilot} have been used for regression tasks. In M5prime, the sum of squared errors (SSE) of the sample output values when the two parts of a split block are fitted by two least-squares solutions is minimized to determine the HP used to split a block. The heuristic tends to minimize the total sum of squared errors. This is done in a way that penalizes the larger block so the solution will tend to have equal numbers of samples on the two sides. An excessively large total SSE could arise by penalizing a larger block which has a lower SSE than its companion. A further question is why should the SSE be the same over the image space anyway. Another problem with this greedy heuristic is its lack of a convergence proof to a globally optimal model tree. 

For the CMT, we use a method that leads to a convergence proof of a trained model tree satisfying a specified maximum error rate. This requires unlimited sample data. We introduce an axis-parallel bounding hyper-rectangle (HR) for every block, enclosing the block's convex shape. The cutting HP always passes through the midpoint of the block, suggesting almost equal numbers of samples remain in the two child blocks.  The product of the width in an axis of the bounding HR and the corresponding least squares coefficient suggests an upper-bound to the change of output in that axis across the block.

We feel that using several variables that cause the largest change of output across the block is a good way to select the direction vector of an HP. Cutting blocks in half this way tends to quickly reduce the range of output values in the child blocks. It also enables quicker elimination of variables of small output range across blocks and may enable us to simplify the least-squares computation. These properties of our cuts help us understand why an oblique model tree could be better than a tree that makes cuts perpendicular to one of the axes. We show that although one can use oblique cuts with all axes participating, making sure training converges imposes limits on the set of axes and the direction vectors used for enough of the cuts during tree construction.

\section{Conditions for convergence}
A pixel number $i$ in an image (after pooling) with $d$ pixels has a certain range of values $\{0, 1, \ldots , w_i\}$, which, for our theory, we replace by the real, closed interval $[0,w_i]$. The Cartesian product of these intervals is a compact HR of dimension $d$ which is the image space. It is this space of all possible vectors of intensities of pixel values which will be partitioned by HPs in a model tree. The image space is huge and only a tiny part is used for any particular task.

\textbf{The task:} Given a $C^1$  function $f$ defined by a set of samples from an image space and a constant $\varepsilon > 0$, approximate the output values with a $C^1$ function to within a root mean square error of $\varepsilon$ using a forest of model trees.

Centered at any point P of the image space, which is the domain of $f$, there is an open, axis-parallel hypercube (HC), where the assumed-continuous partial derivatives of $f$ vary less than any desired amount from their values at P. This implies there is such an HC with a linear fit to the function within $\varepsilon$ error or less. We call this a type A HC and assign it to the point P.  We assign to P another open axis-parallel HC, type B, of half the size in each axis with center at P so that any axis-parallel, type-B-sized HC intersecting any such a B lies entirely within the HC of type A. The covering of the original HR by open HC of size B at all points has, by compactness of HR, a finite sub-cover. For that sub-cover, the sizes of axes of HCs have an upper bound.  The bounding HR of any leaf block of a partition of the image space of lesser size than this bound has a point in common with an element of the sub-cover and so is inside its type-A set and has a satisfactory linear fit to $f$ on it. To prove the algorithm below always converges to a solution, we have to show only that the HR of blocks, if they don't already have a satisfactory fit, will ultimately get split into leaf blocks with all axes smaller than the above bound.

\section{Performing a least-squares fit}

All blocks, including the original image space, are treated in the same way. To save indexing of blocks, the sums below are over all samples $(x_1,\ldots, x_d, y)$ in a block unless otherwise indicated. Based on the set of $N$ samples of a block we compute the least-squares fit to this data and get coefficients $\alpha_i$ for $i \in D = \{1, 2, \dots d\}$ and a constant term $\beta$.  We start with computing the components of the centroid vector of the block: $c_i = 1/N \sum x_i $ and the mean of the $y$ values over all samples of the block: $Y = 1/N \sum y$. The expression to be minimized is

\begin{equation}
  \sum  \{\sum_{i=1}^{d} {\alpha_i (x_i - c_i) + \beta - y \}^2 }
\end{equation}
We take the partial derivatives w.r.t. $\alpha_i$ and $\beta$ and set them to zero to obtain $\beta = Y$ and for all $i \in D$
\begin{equation}
\alpha_i = \frac { N \sum (x_i y) - \sum x_i  \sum y} { N \sum x_i ^ 2 - (\sum x_i)^2 }  
\end{equation}

 You can divide the numerator and the denominator by $N^2$ to get averages, but we keep the above for more efficient computation.  For that purpose, we can append to each input vector the squares  $(x_1^2, x_2^2, \dots x_d^2)$ and the products $(x_1 y, x_2 y, \dots x_d y)$. These extended vectors will save doing some multiplications with every new block and leaf-function while forming the model tree. They don't have to be copied during tree construction, just pointed to when needed for an operation.  One evident  efficiency is that  the sums needed above for the two child blocks add to the value of the parent block. By doing just the sums for one child, the sums for the other child can be obtained by subtraction.

\section{Removing variables from a block}

The RMS error of the samples on the block from the least squares fit is computed.  Here we have to assume that there are enough samples to decide if the fit to $f$ over the block is satisfactory. If it is, the fit becomes the leaf-block function, otherwise, the block is split.

We initially define a hyperplane(HP) using the $\alpha_i, i \in D$, with a constant term set to place the HP so it goes through the midpoint of the axis-parallel bounding HR. Suppose $m_i$ is the midpoint value of axis $i$ of the bounding HR of this block. Let the bounding interval of axis $i$ be $[m_i - h_i, m_i + h_i]$. Then the following holds for the coordinates $x_i$  of any point on any HP through the midpoint:

\begin{equation}
\sum_{i \in D} { \alpha_i(x_i - m_i)} = 0
\end{equation}

The vector with components $\alpha_i$ is normal to the HP in the image space. $|\alpha_i|$ shows how much the output variable y of the fit changes with a change of $x_i$. More importantly, if this is multiplied by the width of the block in direction $i$ to get $\beta_i = |2\alpha_i h_i|$, then we get an upper bound on the magnitude of change that y could undergo across the block, at least according to the linear model. The width of the block in direction $i$ varies as other $x_j$ values change. We avoid calculations involving the geometry of the convex block and simplify by using the bounding HR instead.

It is appropriate to use $|\alpha_i| h_i$ to measure the \emph{importance} of variable $i$ in a block. We can remove any variable $x_i$ less important than some fraction of $\varepsilon$ so that the accumulation of many removals won't disturb the accuracy of the final fit to $f$.

\section{Performing a convolution on the HP coefficients}

The HP coefficients of a model tree can be thought of as laid out in a 2D grid in 1-to-1 correspondence with the grid of image pixel locations. The intensity of a pixel corresponds to the coefficient estimating the linear growth of the output variable due to that intensity. The convolution operation replaces the original coefficients by spreading part of the original value of a coefficient around to the  locations of its neighbors in the grid, with the fraction spread depending on the distance of the neighbor from the original location. We ignore edge effects at image boundaries in this paper. The result of convolution is a new direction vector for an HP and a different function for a leaf. When a leaf of the tree is reached during training, the convoluted coefficients of what would have been a new HP for splitting the block are used, together with the average output value on the block, for the leaf-function. We continue to use the same symbol $\alpha_i$ for convolved coefficients of HPs in what follows.

The motivation for convolution on images is. in effect, to enlarge the training set to include some slightly translated one-pixel parts of images. The goal of doing this on HPs is to avoid having to perform time-consuming convolution on images after deployment. The goal of using convolution on the leaf-function is to correctly handle images that are close to training points in the domain.

\section{A tilt constraint for a splitting HP}

A block is split into two child blocks so that the output values on one side of the HP are generally greater than on the other side. Clearly the split should be in one of the important axes or a combination of them to raise the contrast of output values between the sides. This is our heuristic and we don't need to improve it to get convergence. Taking all the HP coefficients from least squares to split may not create child blocks with smaller bounding HRs. For example, a diagonal split of a square leaves the bounding rectangles of the children equal to that of the parent. Tilt the split so it is more horizontal and you get two pieces with are shorter along one axis. We need some constraints to make sure that the size of the bounding HR decreases often enough during construction of the model tree so that the size in each axis approaches zero. That will assure that our algorithm converges. A child's bounding box is never larger than the parent's.

We first pick the most important axis $k$ and transform the midpoint-centered HP equation so that $\alpha_k (x_k - m_k)$ is on the left side and the other terms are on the right. We set all $x_i$ on the right to $m_i \pm h_i$ at either end of its range inside the bounding box. By choosing the signs, we can maximize or minimize $x_k$ on the left, which is the axis  we want to shorten. That means each term must be of the form $|\alpha_i| h_i$ in the following:

Let $D_k = D - \{ k \}$.

\begin{equation}
	\alpha_k (x_k - m_k) = \pm \sum_{i \in D_k } |\alpha_i|  h_i
\end{equation}.

The important question is how far $x_k$ can deviate from $m_k$. If it deviates by more than $\pm h_k$, then where it cuts the corner is outside the HR. To make sure the cut will shorten the $h_k$ to as small as we need in the children, we take a $\tau \in (0,1)$ and define what we call the \emph{tilt constraint}  as follows: 

\begin{equation}
	\tau |\alpha_k| h_k \ge \sum_{i \in D_k}|\alpha_i| h_i
\end{equation}

If this is satisfied, then we have by symmetry a size constraint on each child that half of the size of the bounding box in axis $k$ is at most $\frac{\sum_{i \in D_k}|\alpha_i| h_i}{|\alpha_k|}$, and we can set this as the value of $h_k$ for the bounding box of each child. Figure 1 shows the HR and the HP cutting it, and any enclosed convex set (not seen), into two pieces, each fitting into a smaller HR along axis $k = 1$.

\begin{figure}[ht]
	\centering
\begin{tikzpicture}[x=1cm,y=1cm,line cap=round,line join=round,scale=0.8,every node/.style={transform shape}]
	
	\tikzset{stroke/.style={draw=black,line width=1.3pt}}
	
	\draw (-3, 2) 
	-- (-3, -4)
	-- (0.8, -4)
	-- (2.6,-2)
	-- (2.6, 4)
	-- (-1.2, 4)
	-- cycle;
	
	\draw (-3, 2)
	-- (0.8, 2)
	-- (2.6, 4);
	
	\draw (-3, -4)
	-- (-1.2, -2)
	-- (2.6,-2);
	
	\draw (-1.2, -2) -- (-1.2, 4);
	\draw (0.8, -4) -- (0.8, 2);
	
	\draw (-3, -3.4)
	-- (-1.2, 2.4)
	-- (2.6, 3.4)
	-- (0.8, -2.2)
	-- cycle;
	
	\draw (-3,-3)
	-- (-1,-3)
	-- (-1,-1)
	-- (0,0);
	
	\draw[<-] (-1,-0.7) -- (-0.65,-0.35);
	\draw[->] (-0.25, 0.05) -- (0,0.3);
	\node at (-0.45,-0.15) {$h_3$};
	
	\draw[<-] (-0.8,-1) -- (-0.8,-1.6);
	\draw[->] (-0.8,-2.1) -- (-0.8,-3);
	\node[xshift=12pt, yshift=5.5pt] at (-0.8,-2) {$|\alpha_3|\,h_3$};
	
	\draw[<-] (-3, -2.8) -- (-2.4, -2.8);
	\draw[->] (-2, -2.8) -- (-1, -2.8);
	\node at (-2.2,-2.8) {$h_2$};
	
	\draw[<->] (-3.2,-3) -- (-3.2, -3.4);
	\node at (-3.8, -3.2) {$|\alpha_2|\,h_2$};
	
	\draw[<-] (0.2, -1) -- (0.2, -2.6);
	\node at (0.2, -2.8) {$h_1$};
	\draw[->] (0.2, -3) -- (0.2, -4);
	
	\draw[->, decorate, decoration={snake, amplitude=0.5mm, segment length=5mm}]
	(3,1.4) -- (2 ,1.4);
	\node at (3.5, 1.4) {HP};
	
	
	\begin{pgfonlayer}{background}
		\node[draw, inner sep=8pt, fit=(current bounding box)] {};
	\end{pgfonlayer}
	
\end{tikzpicture}
\caption{ Illustrating $h_1 > |\alpha_2|\,h_2 +|\alpha_3|\,h_3$};
\end{figure}
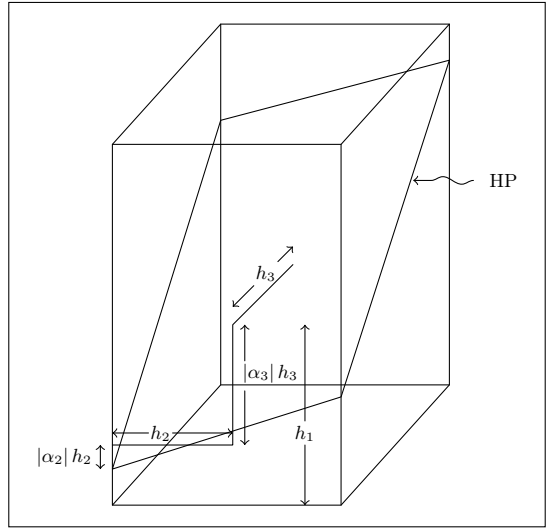

What happens if the cut does not reduce the width of the axis $k$?  We can still make the cut, but it may not lead us to a solution as fast. To reduce the size of the bounding box along axis $k$, we change the direction vector.  We can change the $\alpha_i$s on the right in three ways. This is often called "shrinkage".
\begin{enumerate}
	\item
	 We can set some of them to zero. Some sensor inputs may have almost no effect on the $y$ output over the block as measured by their importance value. One could set the HP coefficient to zero if the importance is, say, less than a fraction of $\varepsilon$ depending on their cumulative effect. On the other hand, one would prefer not to eliminate from the HP coefficients belonging to sensor inputs of high importance. Those are doing precisely what we want them to do: highlight the differences between higher and lower values of $y$ in the two child blocks.
	 \item
	 We can diminish some of them on the right to a fraction of their original value, which changes the direction vector and helps to correct a violation of the tilt constraint.
	 \item
	 We can reduce all the $\alpha_i$ on the right by the same positive factor, which can correct any tilt violation.
\end{enumerate}

 Picking he most important axis to split assures that no axis will be left out of cuts using the tilt constraint. In deciding to set some $\alpha_i$ to zero we may have to consider the hardware used to compute inner products in parallel. The hardware may have a maximum parallelism of N products of a pixel value with an HP coefficient. In that case, we must eliminate all but N variables.  This could be done in order of increasing importance.
 
 We present the following theorem and we leave it up to the reader to deal with tricky questions about the error at HP boundaries, the number of samples to compensate for noise, how to translate this theorem into guidelines for real computer applications and so on. 

\textbf{Theorem}\textit{ Given an unlimited number of random samples of a real $C^1$ function defined on an image space, an error bound $\varepsilon$ and a positive $\tau < 1$ for the tilt constraint, splitting the HPs of a CMT while applying the tilt constraint a sufficient number of times to make all leaf blocks' bounding boxes as small as necessary in all axes leads to an approximation of the ideal function by the CMT to within RMS error $\varepsilon$.}

\section{Weighting a leaf function so the boundaries vanish}

We introduce weight functions over all leaf blocks of the model tree. This is useful only after constructing several trees in a forest when processing a new input image to get the regression output of the forest. 
All HP are defined by an equation setting a certain function $S$ defined on the images of the block to zero. Each HP which initially just separates two child blocks of a block, can be a separator of many pairs of its sub-blocks resulting from further partitioning later in the process. The inner product of the input image with whatever coefficients have been retained in the direction vector of the HP is computed and the constant term is added to get the value of $S$ and the image is triaged to the correct child according to the sign of $S$. Then there are different possibilities. The weight value assigned to the image for this HP can be the minimum or product for all splits from root to a leaf of
\begin{enumerate}
	\item the absolute value of $S$. 
	\item the cubic function $W(|S|) = 3 |S|^2 -  2 |S|^3 $ for $|S| < 1$ and $W(|S|) = 1$ for $S \ge 1$.
	\item a more complicated function that takes account of the size of adjacent blocks and has value zero at all block boundaries and 1 at the midpoint of all blocks.
\end{enumerate}

  As the image passes down the tree towards a leaf, the minimum (or the product) of the $W(|S|)$ so far encountered with that of the current HP is taken, to be passed along down the tree until a leaf is reached. If the product is taken, then the cubic function gives rise to a $C^1$ weight on the block's leaf function. Taking the minimum gives a continuous weight. Using a sigmoidal function as in neural nets, it would be possible to get a $C^{\infty}$ weight function.

\section{Using a forest to get an accurate, smooth fit}

We create a forest of model trees as above, whereby they are either newly trained trees or they are shifted copies of the first tree.  In the case of shifts, the trees are shifted by the same amount from the previously shifted tree in all axes of the domain. Thus each tree has blocks that are constructed using, in part, different training samples.  This has the desirable effect of making the weighted average of their leaf functions more accurately fit $f$. All boundaries have been shifted so corresponding boundaries of two trees of the forest have no common zeros. However, there will likely be many places where boundaries of different origin intersect in lower dimensional spaces leading to zero weights in several or all trees. As long as there is one tree with a positive weight, the weighted average is defined.

We need a practical solution to prevent the arithmetic mean of the weighted tree outputs being indeterminate when a division by zero is called for. We pick a positive number $\mu$ near zero. Consider the function $W(w) = 0$ for $w > \mu$ and $ W(w) = \mu - w$ otherwise, and a "helper tree" added to the forest which is just a constant value, say the mean of the values of all outputs of samples we use for training the model trees. The constant has weight $W(w)$ in the forest's weighted average. This will prevent crashes and won't change the error of approximation much if we choose a small enough number for $\mu$. In theory, the $C^1$ property of the weighted average is also preserved.

 Perhaps the most important result of smoothing is that output functions of forests will not show a kink or discontinuity, despite the discontinuities of model tree outputs.

\section{Distilling a neural net into a CMT}

A back-propagation neural net learns a function that inevitably has no samples in large parts of a high-dimensional image space.  Nevertheless, the over-all result can be extremely useful. However, if there are inputs given in an area where samples are lacking, an adverse event could result. Weight decay, a global restriction, may be used to try to prevent unwanted excursions of the net output, but it has to allow for required  value changes. A test set provides single values at places in the domain. This section suggests that obtaining a piecewise-linear approximation of the neural net output with a known maximum error can tell us what the neural net output can be everywhere in the the part of the domain of interest to us.

A CMT needs enough samples to determine the HPs and leaf-functions even in the presence of noise. Often, there will not be enough data to reach a desired low level of RMS error. Distilling the output of a deep neural net is one application where the CMT has an opportunity to shine, that is to use samples created from a deep neural net's training set and its computed outputs \cite{frosst2017distilling,hinton2015distilling} to train a model tree. If the neural net is a CNN, the tree does not need to do the convolution if it has already been done and stored. The training (and test) samples input to the neural net or CNN can be augmented by sampling in regions where a deployment-time input could be presented. On those we we have to do the convolution. It might be preferable always to do the convolution in the CMT.  We need many samples just to cover the tiny part of the image space needed for solving the desired task. According to our convergence theorem, we can hope to create an arbitrarily close approximation of the net output in a useful part of the domain this way. The assumption that ideal function is $C^1$ may not be satisfied because of kinks in the CNN caused by rectified units etc., but a forest of CMTs will produce a smooth function regardless of this.

The leaf-block functions of a single tree demonstrate the behavior of the CNN output, and one can check the all the coefficients of the tree for unexpectedly large slopes that might indicate some unwanted peaks or troughs that were not caught by using a test set. This may happen more often than we are aware of since image spaces are so huge.

\section{Thoughts about special hardware}
Below are reasons why we believe it would be easy to design special hardware for certain operations of a model tree after deployment that would make its use fast with low power consumption:
\begin{enumerate}
	\item not all of the tree has to be loaded at once, a single HP or leaf function is all that needs to be executed at one time. This is an inner product done in parallel for speed.
	\item each HP makes a binary choice of the next HP or leaf function to be evaluated, which allows for pre-computing a small number 3, 7, 15 etc. of the following inner products before it is known which ones will be needed. 
	\item the same 3, 7, 15 .. hardware units are all that is required for the whole tree, since they can be reloaded and re-used. This leaves room for having many hardware units much smaller than entire model trees to perform different tasks such as evaluating forests for a needed output or doing entirely different tasks at the same time.
	\item unlike a back-propagation neural net, the model tree hardware does one task: compute an inner product from the inputs and data stored in memory. Other tasks, e.g. computing weights or averages, can be done by the host computer. The variety of simultaneously active linear units in a back-propagation neural net, rectified units, squashing functions and so on, is not required.
\end{enumerate}

\section{Distortions}

  This section of the paper does not depend on the training technique used for a tree. Only the one-to-one correspondences of the locations of pixels, HP coefficients and leaf coefficients are needed. For the purposes of exposition, we are going to use an imaginary project to learn about identification and flight control in birds of prey like hawks, owls and peregrine falcons, say with a view to building a robotic model. 

Imagine you have an elevated blind on the edge of a field on the outskirts of a city, and you are capturing high-resolution videos of birds as they patrol in the air, dive, and land or attack their prey: mice, rats etc. The video images are processed by placing rectangles around moving objects, from which the background can often be stripped away and replaced by a single value since it is motionless. The intensity and contrast can be standardized.

Next,some images very similar to a particular one could be grouped to form a training set. Say when the birds' bodies are nearly horizontal. Measured numbers would be assigned to each image, which could be anything from the human-determined species, to age, wingspan etc. These would be used to train a model tree, maybe distilled from a deep neural network. The domain of usefulness would be enlarged by convolution applied to the HP coefficients. Next, we can enlarge the set of bird configurations. We start with one very simple situation where the distortion could be just one wing moved in a horizontal direction by a number of grid units like Figure 2 and the pixels only need to be permuted.

\newcommand{\Eye}{
	\draw (-3.40,-1.72) -- (-3.15,-1.5);
	\draw[shift={(-3.275,-1.61)}, rotate=42] (0.1665,0) arc (0:-180:0.1665);
}

\newcommand{\BeakInnerLine}{
	\draw (-3.64,-2)
	-- (-3.5, -2.2)
	.. controls (-3.35, -2.1) .. (-3.2, -2.05)
	-- (-3.3,-2.2);
}

\newcommand{\Body}{
	\draw (-3.65,-2) .. controls (-3.67,-1.10) and (-2.25,-1.15) .. (-1.8,-0.6);
	\draw (-1.8,-0.6) .. controls (-1.83,-0.75) .. (-1.85,-0.78);
	
	\draw (-3.3,-2.2)
	.. controls (-3.42, -2.4) .. (-3.4, -2.6)
	.. controls (-3.65, -2.4) .. (-3.64,-2);
	
	\draw (-3.3,-2.2) .. controls (-1.8,-1.6) and (-1.8, -2.7) .. (-0.7,-2.69);
	\draw (0.1, -2.7) .. controls (0.6, -2.7) .. (1.2,-2.4);
}

\newcommand{\LeftLeg}{
	\draw
	(-1.2, -2.6)
	.. controls (-0.75, -3.4) .. (-1.15, -4)
	.. controls (-1.75, -4) and (-2, -4.2) .. (-1.8, -4.4)
	.. controls (-1.75, -4.2) .. (-1.6 , -4.15)
	.. controls (-1.7, -4.4) .. (-1.6,-4.6)
	.. controls (-1.5, -4.2) .. (-0.8, -4.1)
	-- (-1 ,-4.05)
	.. controls (-0.2, -3.45) .. (-0.7, -2.7);
}

\newcommand{\RightLeg}{
	\draw
	(-0.4, -3.5)
	-- (-0.7, -4.2)
	.. controls (-1.2, -4.3) .. (-1.2, -4.6)
	.. controls (-1.1, -4.35) .. (-0.9, -4.35)
	.. controls (-1.15, -4.7) .. (-0.9, -5)
	.. controls (-1, -4.65) .. (-0.6, -4.3)
	.. controls (-0.4, -4.3) .. (-0.2, -4.6)
	.. controls (-0.2, -4.3) .. (-0.45, -4.2)
	-- (-0.1, -3.7)
	.. controls (0, -3.8) .. (0.1, -4)
	.. controls (0.2, -3.6) .. (0.18, -3.3)
	.. controls (0.3, -3.4) .. (0.4, -3.6)
	.. controls (0.3, -3) .. (0.1, -2.7);
}

\newcommand{\LeftWing}{
	\draw
	(-2.5, -1)
	.. controls (-2.6, -0.1) and (-4, 0) .. (-4.2, 0.3)
	.. controls (-4.7, 0.8) and (-4.5, 1.6) .. (-5.1, 2.1);
	
	\draw (-5.1, 2.1)
	.. controls (-4, 1.7) .. (-3,1)
	.. controls (-2.1, 0.7) .. (-1.4, 0);
}

\newcommand{\RightWing}{
	\draw (-1.85,-0.78) .. controls (-0.8, 0.6) and (-1.6, 1.4) .. (0, 2.4);
	\draw (0, 2.4) .. controls (1.425, 3.4) .. (2.85, 5.4);
	
	\draw (0,-1.25) .. controls (0.7,-1.1) and (1.4,-1.4) .. (1.6,0.8);
	\draw (1.6,0.8) .. controls (3.6, 3.2) .. (2.85, 5.4);
}

\newcommand{\Tail}{
	\draw (1.2,-2.4) .. controls (2.2, -2.8) .. (3.2,-2.8);
	\draw (1.8,-2.2) .. controls (2.5, -2.6) .. (3.2,-2.8);
	
	\draw (1.8,-2.2) .. controls (3.8, -2.8) .. (4.45, -2.4);
	\draw (0, -1.25) .. controls (2.225, -2.15) .. (4.45, -2.4);
}

\newcommand{\WingDetails}{
	\draw (-1.8,-1)
	.. controls (-0.4, 0.4) and (-1.4, 1) .. (0.4, 2.4)
	.. controls (1.7, 3.4) .. (2.85, 5.4);
	
	\draw (-1.8,-1) .. controls (-0.5, -0.6) and (-1.3, 1.4) .. (1.2, 2.8)
	.. controls (0.7, 1.2) .. (0.3, 1)
	.. controls (0.2,-1) and (-0.6, -0.4) .. (-1.7, -1.1);
	
	\draw (-5.1, 2.1)
	.. controls (-4, 1.4) and (-4.67, 0.4) .. (-3.2, 0)
	.. controls (-2.95, -0.1) .. (-2.6, -0.4)
	.. controls (-3, 0.2) and (-4.2, 0.3) .. (-4.1, 1)
	.. controls (-3.6, 0.9) .. (-3.2, 0.5)
	.. controls (-2.6, 0.4) .. (-1.9, -0.5);
}

\newcommand{\Feathers}{
	\draw (1.3, 2.8)
	-- (1.17, 2.4)
	.. controls (2.1, 3.2) .. (3, 5)
	.. controls (2.1, 3.6) .. (1.3, 2.8)
	-- cycle;
	
	\draw (1.16, 2.3)
	-- (1.05, 2)
	.. controls (2.2, 2.6) .. (3.2, 4.2)
	.. controls (2.2, 3) .. (1.16, 2.3)
	-- cycle;
	
	\draw (1.03, 1.9)
	-- (0.92, 1.6)
	.. controls (2.2, 2) .. (3.3, 3.3)
	.. controls (2.2, 2.3) .. (1.03, 1.9)
	-- cycle;
	
	\draw (0.9, 1.5)
	-- (0.75, 1.2)
	.. controls (1.5, 1.3) .. (2.4, 1.8)
	.. controls (1.5, 1.5) .. (0.9, 1.5)
	-- cycle;
	
	\draw (0.7, 1.1)
	-- (0.4, 0.9)
	.. controls (1.2, 0.9) .. (1.92, 1.2)
	.. controls (1.4, 1.1) .. (0.7, 1.1)
	-- cycle;
	
	\draw (0.4, 0.8)
	-- (0.37, 0.5)
	.. controls (0.8, 0.55) .. (1.6, 0.8)
	.. controls (0.8, 0.7) .. (0.4, 0.8)
	-- cycle;
	
	\draw (0.35, 0.4)
	-- (0.26, 0.1)
	.. controls (0.8, 0.15) .. (1.57, 0.4)
	.. controls (0.8, 0.35) .. (0.35, 0.4)
	-- cycle;
	
	\draw (0.25, 0)
	-- (0.07, -0.3)
	.. controls (0.6, -0.35) .. (1.47, 0)
	.. controls (0.7, -0.1) .. (0.25, 0)
	-- cycle;
	
	\draw (0, -0.4)
	-- (-0.35, -0.67)
	.. controls (0.4, -0.7) .. (1.4, -0.4)
	.. controls (0.4, -0.5) .. (0, -0.4)
	-- cycle;
	
	\draw (-0.5, -0.75)
	-- (-0.9, -0.85)
	.. controls (0.4, -1.1) .. (1.2, -0.8)
	.. controls (0.4, -0.9) .. (-0.5, -0.75)
	-- cycle;
	
	\draw (-3.9, 1)
	-- (-4.1, 1.05)
	.. controls (-4.2, 1.5) .. (-4.4, 1.8)
	.. controls (-4.1, 1.5) .. (-3.9, 1)
	-- cycle;
	
	\draw (-3.8, 1)
	-- (-3.6, 0.94)
	.. controls (-3.62,1.3) .. (-3.9, 1.6)
	.. controls (-3.78, 1.3) .. (-3.8, 1)
	-- cycle;
	
	\draw (-3.55, 0.89)
	-- (-3.37,0.77)
	.. controls (-3.35, 1) .. (-3.5, 1.35)
	.. controls (-3.47, 1) .. (-3.55, 0.89)
	-- cycle;
	
	\draw (-3.3, 0.7)
	-- (-3.16, 0.57)
	.. controls (-3.1, 0.8) .. (-3.2, 1.12)
	.. controls (-3.2, 0.8) .. (-3.3, 0.7)
	-- cycle;
	
	\draw (-3, 0.53)
	-- (-2.8, 0.47)
	.. controls (-2.85,0.7) .. (-3, 1)
	.. controls (-2.95, 0.7) .. (-3, 0.53)
	-- cycle;
	
	\draw (-2.7, 0.45)
	-- (-2.5, 0.3)
	.. controls (-2.57,0.6) .. (-2.7, 0.9)
	.. controls (-2.67, 0.6) .. (-2.7, 0.45)
	-- cycle;
	
	\draw (-2.42, 0.22)
	-- (-2.27, 0.05)
	.. controls (-2.25, 0.4) .. (-2.45, 0.81)
	.. controls (-2.4, 0.4) .. (-2.42, 0.22)
	-- cycle;
	
	\draw (-2.2, 0)
	-- (-2.05, -0.2)
	.. controls (-2.05, 0.2) .. (-2.15, 0.66)
	.. controls (-2.15, 0.2) .. (-2.2, 0)
	-- cycle;
	
	\draw (-2.03, -0.25)
	-- (-1.92, -0.4)
	.. controls (-1.8, 0) .. (-1.9, 0.5)
	.. controls (-1.9, 0) .. (-2.03, -0.25)
	-- cycle;
	
	\draw (-1.87, -0.45)
	-- (-1.77, -0.56)
	.. controls (-1.6 ,-0.2) .. (-1.7, 0.3)
	.. controls (-1.75, -0.2) .. (-1.87, -0.45)
	-- cycle;
}

\newcommand{\RightWingOutline}{
	\draw
	(-1.85,-0.78)
	.. controls (-0.8, 0.6) and (-1.6, 1.4) .. (0, 2.4)
	.. controls (1.425, 3.4) .. (2.85, 5.4)
	.. controls (3.6, 3.2) .. (1.6, 0.8)
	.. controls (1.4,-1.4) and (0.7,-1.1) .. (0,-1.25);
}

\newcommand{\RightWingOutlineDash}{
	\draw
	(-1.85,-0.78)
	.. controls (-0.8, 0.6) and (-1.6, 1.4) .. (0, 2.4)
	.. controls (1.425, 3.4) .. (2.85, 5.4)
	.. controls (3.6, 3.2) .. (1.6, 0.8)
	.. controls (1.4,-1.4) and (0.7,-1.1) .. (0,-1.5);
}

\newcommand{\LefttWingOutlineDash}{
	\draw
	(-2.5, -1)
	.. controls (-2.6, -0.1) and (-4, 0) .. (-4.2, 0.3)
	.. controls (-4.7, 0.8) and (-4.5, 1.6) .. (-5.1, 2.1);
	
	\draw (-5.1, 2.1)
	.. controls (-4, 1.7) .. (-3,1)
	.. controls (-2.1, 0.7) .. (-1.7, 0.3);
}

\newcommand{\LegsDash}{
	\draw
	(-1.2, -2.8)
	.. controls (-0.75, -3.4) .. (-1.15, -4)
	.. controls (-1.75, -4) and (-2, -4.2) .. (-1.8, -4.4)
	.. controls (-1.75, -4.2) .. (-1.6 , -4.15)
	.. controls (-1.7, -4.4) .. (-1.6,-4.6)
	.. controls (-1.5, -4.2) .. (-0.8, -4.1)
	-- (-1 ,-4.05)
	.. controls (-0.2, -3.45) .. (-0.7, -2.7)
	(-0.4, -3.5)
	-- (-0.7, -4.2)
	.. controls (-1.2, -4.3) .. (-1.2, -4.6)
	.. controls (-1.1, -4.35) .. (-0.9, -4.35)
	.. controls (-1.15, -4.7) .. (-0.9, -5)
	.. controls (-1, -4.65) .. (-0.6, -4.3)
	.. controls (-0.4, -4.3) .. (-0.2, -4.6)
	.. controls (-0.2, -4.3) .. (-0.45, -4.2)
	-- (-0.1, -3.7)
	.. controls (0, -3.8) .. (0.1, -4)
	.. controls (0.2, -3.6) .. (0.18, -3.3)
	.. controls (0.3, -3.4) .. (0.4, -3.6)
	.. controls (0.3, -3) .. (0.1, -2.7);
}

\newcommand{\BodyTailDash}{
	\draw
	(-3.65,-2) .. controls (-3.67,-1.10) and (-2.25,-1.15) .. (-1.8,-0.6)
	(-1.8,-0.6) .. controls (-1.83,-0.75) .. (-1.85,-0.78)
	(-3.3,-2.2)
	.. controls (-3.42, -2.4) .. (-3.4, -2.6)
	.. controls (-3.65, -2.4) .. (-3.64,-2)
	(-3.3,-2.2) .. controls (-1.8,-1.6) and (-1.8, -2.7) .. (-0.7,-2.69)
	(0.1, -2.7) .. controls (0.6, -2.7) .. (1.2,-2.4)
	(1.2,-2.4) .. controls (2.2, -2.8) .. (3.2,-2.8)
	(1.8,-2.2) .. controls (2.5, -2.6) .. (3.2,-2.8)
	(1.8,-2.2) .. controls (3.8, -2.8) .. (4.45, -2.4)
	(0, -1.25) .. controls (2.225, -2.15) .. (4.45, -2.4);
}

\begin{figure}[ht]
	\centering
	\setlength{\fboxsep}{16pt}
	\fbox{%
		
		\begin{tikzpicture}[stroke, x=1cm, y=1cm, line cap=round, line join=round, scale=0.8,every node/.style={transform shape}]]
			\def\FrameHalfW{4.5cm}
			\def\FrameHalfH{5cm}
			\def\HawkScale{0.87}
			\coordinate (leftframe) at (0,0.5);
			\coordinate (rightframe) at (9.8,0.5);
			
			
			\begin{scope}[scale=\HawkScale]
				\Body
				\Eye
				\BeakInnerLine
				\Tail
				\RightWing
				\LeftWing
				\LeftLeg
				\RightLeg
				\WingDetails
				\Feathers
			\end{scope}
			
			\begin{scope}[xshift=10cm]
				\begin{scope}[scale=\HawkScale]
					\Body
					\Tail
					\RightWing
					\LeftWing
					\LeftLeg
					\RightLeg
				\end{scope}
			\end{scope}
			
			\begin{scope}[xshift=9.8cm, yshift= -0.05cm]
				\begin{scope}[scale=\HawkScale, dashed, dash pattern=on 6pt off 4pt, line cap=round]
					
					\begin{scope}[rotate around={4:(-1.85,-0.78)}, shift={(0.15,0.15)}]
						\RightWingOutlineDash
					\end{scope}
					
					
				\end{scope}
			\end{scope}
			
			\draw ($(leftframe)+(-\FrameHalfW,-\FrameHalfH)$) rectangle
			($(leftframe)+(\FrameHalfW,\FrameHalfH)$);
			\draw ($(rightframe)+(-\FrameHalfW,-\FrameHalfH)$) rectangle
			($(rightframe)+(\FrameHalfW,\FrameHalfH)$);
			\node[align=center] at ($(leftframe)+(0,\FrameHalfH-0.4cm)$)
			{Sensor values in a picture};
			\node[align=center] at ($(rightframe)+(0,\FrameHalfH-0.4cm)$)
			{HP coefficients in 1-to-1 correspondence};
			
		\end{tikzpicture}
	}
	\caption{The dashed outline shows where the left wing, close to us, has moved. Coefficients inside the dotted lines and in front of original outline of the left wing are moved, randomly placed, to the space behind the new wing position. This assumes that moved parts of the bird don't change much in intensity and that the background pixels all have a nearly constant value. This keeps inner products almost unchanged from the original position }
	\label{fig:mytikz}
\end{figure}
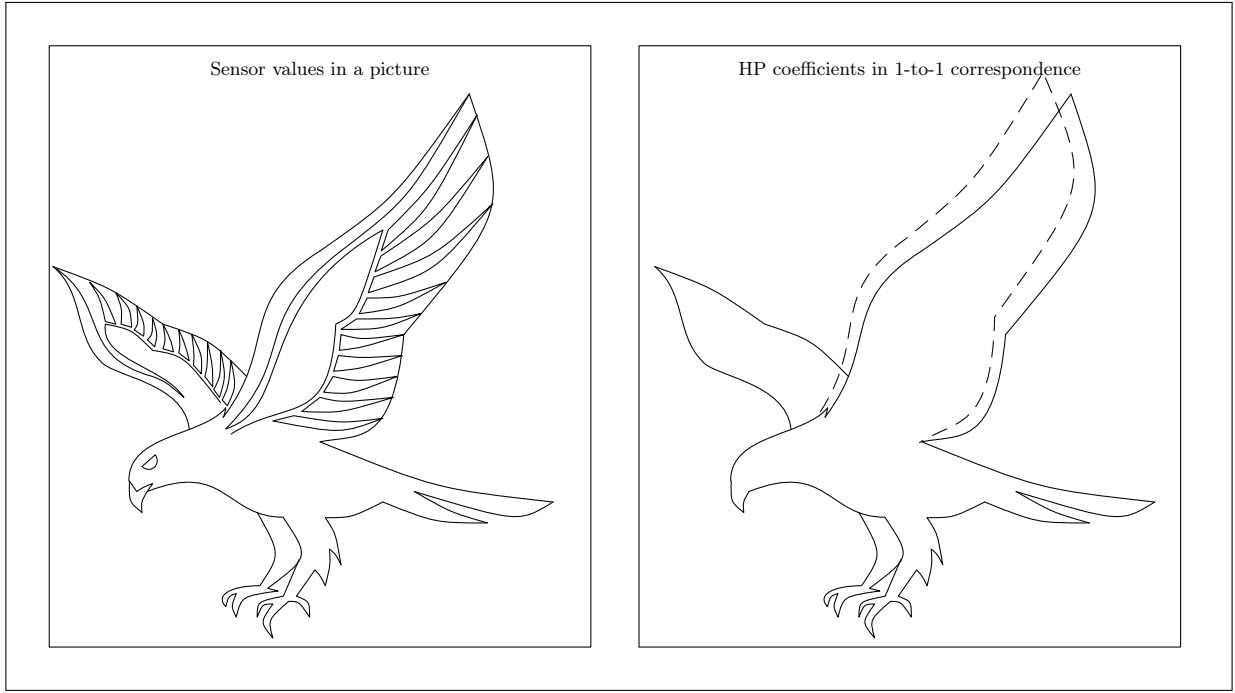

Now suppose we want to deal with images of objects in a much larger set of configurations. Dealing with all possible rotations is easy if we use a circular grid.  The "control panel" values of HP coefficients from the original training are simply rotated in position by a discrete set of angles about a central point and each angle gives rise to a new model tree. Rotations between the discrete set are taken care of by the convolution.

For a rectangular grid, moving a coefficient geometrically like a part of the image moves usually requires distributing the coefficient's value among the four values surrounding the destination. Under the assumption that the picture intensity or color remains the same at the destination, this doesn't change the inner products. Taking the grid cells as 1 unit wide and one unit tall, one has the fractional parts of the destination horizontally and vertically $H$ and $V$ at the bottom left. In one dimension, if the division is into intervals of size $H$ and $1 - H$ on the left and right respectively, then the fraction of the weight going to the $H$ part is $1-H$ and \textit{vice versa}. For the vertical fraction $V$ touching the bottom line the are values are distributed similarly. In $2D$ we take the products, for example the lower right corner would ge a fraction $H(1 - V)$ of the value of the original pixel. The center of gravity is at the destination. 

If we want to process images of half the size, several pixel values would have to be merged to form one. The details of all useful possibilities, including perspective transformations and bending objects have to be worked out based on the above.

Suppose we don't know what kind of object the system is seeing in a certain position or its orientation. Rapid loading and execution of many forests containing possible object types and orientations could be executed with class-conditional probabilities as the regression function outputs.

Some experiments that used several of the ideas in this paper were carried out on distortion \cite {LiArmXu_Robust}.  MNIST digits under rotations up to ±60° were used to generate rotation-invariant targets: a perimeter-based regression target and a digit identity  target via one-vs-rest regression. A deployment-time orientation search selects an input rotation from a small discrete angle set by maximizing a forest-level confidence proxy (with model parameters fixed). Ablations quantify the effects of convolutional smoothing, the tilt constraint, and importance-based pruning, including accuracy and computation time trade-offs.

\section{Conclusion}

This is an "ideas" paper. We have not investigated whether some combination of the ideas above for CMTs and  ideas from other training algorithms for model trees could lead to an improved CMT algorithm. The work done in \cite {LiArmXu_Robust} shows that the idea of distortions works, and that it is possible to find approximately how much an image of a known class is rotated. Given how well this is accomplished for an arbitrary, partially obscured perspective view of an object by the human eye and brain, we see that much remains to be done. This paper joins the capabilities of neural nets or CNNs and CMTs to advance computer vision. It poses a challenge to researchers to think about further development of CMTs.

\vspace*{1em}
\textbf{Acknowledgement}
The drawings for this paper were created using LaTeX and tikZ by Keya Malhotra.
\printbibliography
\end{document}